%% file: main.tex
\documentclass{article}
\usepackage{ijcai22}
\usepackage{comment}
\usepackage{times}

\usepackage{soul}
\usepackage{url}
\usepackage[hidelinks]{hyperref}
\usepackage[utf8]{inputenc}
\usepackage[small]{caption}
\usepackage{graphicx}
\usepackage{amsmath}
\usepackage{booktabs}
\usepackage{stfloats}
\usepackage[ruled,linesnumbered]{algorithm2e}
\usepackage{amssymb}
\usepackage{multirow}
\usepackage{tikz}
\usetikzlibrary{positioning, patterns, snakes}

\title{Towards Machine Learning for Placement and Routing in Chip Design: a Methodological Overview}
\author{ 
Junchi Yan$^1$\footnote{Correspondence author.}\and
Xianglong Lyu$^1$\and
Ruoyu Cheng$^1$\and
Yibo Lin$^{2*}$\\
\affiliations
$^1$Department of CSE, and MoE Key Lab of Artificial Intelligence, Shanghai Jiao Tong University\\
$^2$Department of EECS, Peking University
\emails  
\{yanjunchi,kyle\_lyu,roy\_account\}@sjtu.edu.cn \quad yibolin@pku.edu.cn
}
\begin{document}

\maketitle


\begin{abstract}
Placement and routing are two indispensable and challenging (NP-hard) tasks in modern chip design flows. Compared with traditional solvers using heuristics or expert-well-designed algorithms, machine learning has shown promising prospects by its data-driven nature, which can be of less reliance on knowledge and priors, and potentially more scalable by its advanced computational paradigms (e.g. deep networks with GPU acceleration). This survey starts with the introduction of basics of placement and routing, with a brief description on classic learning-free solvers. Then we present detailed review on recent advance in machine learning for placement and routing. Finally we discuss the challenges and opportunities for future research. 

\end{abstract}

\section{Introduction}
The scale of integrated circuits (ICs) has increased dramatically, posing a challenge to the scalability of existing Electronic Design Automation (EDA) techniques and technologies. The increasing circuit density incurs additional issues for very large scale integration (VLSI) placers and routers as the feature size of modern VLSI design continues to drop and on-chip connectivity gets increasingly sophisticated. Due to increased on-chip connectivity, concentrated needs, and restricted resources, modern designs are prone to congestion issues and wire-length minimization, which has become a critical task at every stage of the design process. Accordingly, placement and routing -- the elements of the design cycle physically arranges the locations and the courses of nets -- becomes more crucial in modern VLSI. 
 
 Placement and routing are two of the most critical but time-consuming steps of the chip design process.
 Placement assigns various logic components like logic gates and functional blocks into discrete sites in the physical layout of a chip, and routing finishes the physical wiring of interconnections. 
 The goal of placement and routing is to optimize power, performance, and area (PPA) metrics within constraints, e.g., placement density and routing congestion. 
 
 Placement can be regarded as a much more complicated variation of 2D bin packing problem with geometrical constraints. The latter is known to be NP-hard~\cite{hartmanis1982computers}. The objective of placement is correlated to both the logical interconnection of the circuit design and the geometrical locations of logical components. 
 As the quality of a placement solution cannot be accurately evaluated until routing, resulting in a long feedback loop in the design flow, modern placement needs to reduce routing congestion and improve routability in early stage. 
 %
 %
 Routing has been proven to be an NP-hard problem~\cite{hartmanis1982computers}, even in the simplest case with only a few two-pin networks. 
 Given the placement solution, it connects the pins of millions of networks with physical wires subjecting to the limited routing resources and complicated geometrical design rules. 
 Routing is tightly coupled with placement and an excellent placement solution leads to better chip area utilization, timing performance, and routability. 

There are also surveys on routing and placement. \cite{huang2021machine} summaries a comprehensive review of existing ML studies for EDA field, most of which belongs to the four categories: decision making in traditional methods, performance prediction, black-box optimization, and automated design, ordered by increasing degree of automation. Although it covers various stages in the EDA flow, it provides shallow analysis for specific tasks, placement and routing, compared to ours. \cite{hamolia2021survey} introduces the ISPD 2015 dataset for comparison between classic methods and ML algorithms. \cite{rapp2021mlcad} categorizes how ML may be used and is used for design-time/run-time optimization and exploration strategies of ICs, along with trends in the employed ML algorithms. \cite{markov2015progress} reviews the history of placement research and the progress leading up to the state of the art. However, they only consider classic placer in the past 50 years, which is complementary to our survey which focuses on learning-based placer in recent years.

\section{Problem Background and Classic Solvers}
This section starts with the problem formulation and mainstream solvers which are traditionally learning-free. We then introduce recent applicable learning techniques. 

\input{flow.tex}

\subsection{Placement Problem}
Placement can be performed at different levels. In general, the global placement involves macro placement and standard cell placement, as shown in Figure~\ref{fig:flow}. The detailed placement includes legalization, wirelength and routability refinement.

\textbf{Global Placement.}
It is one of the most crucial but time-consuming steps in the chip design process, which can be cast as a constrained optimization problem. It assigns exact locations for various components of a netlist including macros and standard cells within the chip layout. Standard cells are basic logic cells, i.e., logic gates, and macros are pre-designed IP blocks, e.g., SRAMs. A good placement leads to better chip area utilization, timing performance, and routability, while inferior placement assignment will affect the chip’s performance and even make it nonmanufacturable.

Formally, the input of global placement is a netlist that can be represented by hypergraph $H = (V, E)$, where $V = \{v_1, v_2, \cdots, v_n\}$ denotes set of nodes (cells), and $E = \{e_1, e_2, \cdots, e_m\}$ denotes set of hyperedges (nets). Hyperedge $e_i \in E$ is a subset of nodes. We seek to determine locations of macros and standard cells $(x_i, y_i)$ which are further combined into two vectors $\vec{x}=\left(x_{1}, \ldots, x_{n}\right)$ and $\vec{y}=\left(y_{1}, \ldots, y_{n}\right)$. 
Wirelength minimization is one of the main objectives for placement. $\text{HPWL}_{N}(\vec{x}, \vec{y})$ is usually adopted to approximate wirelength during placement as:
$$
HPWL=\sum_{e_{i} \in E}\left(\max _{v_{j} \in e_{i}} x_{j}-\min _{v_{j} \in e_{i}} x_{j}+\max _{v_{j} \in e_{i}} y_{j}-\min _{v_{j} \in e_{i}} y_{j}\right)
$$

\textbf{Detailed Placement.}
Solution from global placement is often illegal: cells may overlap or occupy illegal sites, e.g. between placement rows. This is because global placers are unaware of these constraints. Detailed placement performs legalization and quality refinement. The legalization removes overlaps between cells and snaps them to the sites of rows with minimum adverse impact on placement quality. The refinement then takes a legal placement and further improves objectives wirelength and routability by locally moving cells. 

\subsection{Routing Problem}
 Routing often involves two cases: global and detailed routing.
 
 \textbf{Global Routing} connects wires with metal resources on a grid graph $G(V, E)$ representing the physical layout of a chip. 
 Essentially, the physical layout is divided into rectangular areas with each area corresponding to a global routing cell (G-Cell), denoting each vertex $v_i \in V$. 
Each edge $e_{ij} \in E$ represents the joint boundary between abutting G-Cells $v_i$ and $v_j$. 
The capacity for an grid edge $e$, $c_e$, is defined as the maximum number of wires that can cross the grid edge. The usage $u_e$ is defined as the actual number of wires crossing the grid edge. The overflow $o_e$ is defined as $max(0, u_e - c_e)$. 
When multiple metal layers are adopted for routing, a vertical dimension is introduced to the grid graph, where abutting metal layers are connected through vias in the 3D grid graph. 
 
 \textbf{Detailed Routing}
  works on a fine-grained 3D grid graph compared with global routing, considering complicated design rules. 
Each grid edge provides unit routing resource such that a detailed routing path on the grid does not violate the design rules like minimum metal width and spacing. 
Detailed routing also needs to consider preferred routing directions, where abutting routing layers prefer perpendicularly routing directions. 
Consecutive grid edges following the preferred direction (e.g., in X or Y direction) of their corresponding layer are described as a routing track. 
Typical detailed routing algorithms take global routing segments as guidance.
Each global routing segment provides rough hints for the topologies of nets, and detailed routing needs to finalize the actual routing paths on the fine-grained grid graph. 
Nets unable to be routed without conflicts on the grid graph can cause design rule violations, which is a widely-used metric to describe the quality of detailed routing solutions or how congested a design is. 
 
 \subsection{Classic Placement Solvers}
 The history of VLSI placement can trace back to the 1960s~\cite{breuer1977class,fiduccia1982linear}, when partition-based methods adopted the idea of divide-and-conquer: netlist and chip layout are partitioned recursively until the sublist can be solved by optimal solvers. This hierarchical structure makes them fast to execute and natural to extend for larger netlists at the cost of solution quality, since each sub-problem is solved independently. Multi-level partitioning methods based on Fiduccia–Mattheyses heuristics~\cite{agnihotri2005recursive,caldwell2000optimal} were developed afterwards. Analytical placers appeared in the early 1980s, but were eclipsed by annealing methods~\cite{kirkpatrick1983optimization} inspired from annealing in metallurgy that involves heating and controlled cooling for optimal crystalline surfaces. In practice, simulated annealing (SA) optimizes a given placement solution by random perturbations with actions such as shifting, swapping and rotation of macros~\cite{ho2004orthogonal,shunmugathammal2020novel}. Although SA is flexible and able to find the global optimum, it is time-consuming and hard to deal with the ever-increasing scale of circuit. Later, analytical techniques have matured including force-directed methods~\cite{spindler2008kraftwerk2} and non-linear optimizers~\cite{chen2008ntuplace3,kahng2005implementation}. In comparison, the quadratic methods are  efficient but show relatively worse performance, while non-linear optimization approximates the cost functions more smoothly with the cost of higher complexity. Recently, however, modern analytical placers e.g. ePlace~\cite{lu2015eplace} and RePlAce~\cite{cheng2018replace} introduce electrostatics-based global-smooth density cost function and Nesterov’s method nonlinear optimizer that achieve superior performance on public benchmarks. They formulate each node of the netlist as positively charged particle. Nodes are adjusted by their repulsive force and the density function corresponds to system’s potential energy. These analytical methods update positions of cells in gradient based optimization scheme and generally can handle millions of standard cells by parallelization on multi-threaded CPUs using partitioning to reduce the run time.
 
 With increasing design complexity, optimizing traditional placement metrics (HPWL) alone is insufficient in practice~\cite{alpert2010makes}. Therefore, routability-driven placers are proposed to emphasize on routing failures. SimPLR~\cite{kim2011simplr} develops lookahead routing to temporarily bloat cells, modulate target aspect ratio of cell placement areas, and modify anchor positions during quadratic placement. NTUPlace4~\cite{hsu2011routability} adopts probabilistic congestion estimation when modeling pin density. There are also other optimization directions such as timing-driven placement for optimizing circuit delay~\cite{riess1995speed,luo2006new} and IC power optimization for circuit power~\cite{cheon2005power,lee2012obstacle}. Another trend is the increasingly extensive use of intellectual property (IP) modules and pre-designed macro blocks. As a result, mixed-size placement tools are becoming indispensable for physical design. \cite{taghavi2005dragon2005} proposes a hierarchical method to place large scale mixed size designs that may contain thousand of macro blocks and millions of standard cells, based on min-cut partitioning and simulated annealing which are both aware of large macro cells. \cite{chen2008constraint} designs a novel constraint graph-based macro placement algorithm that removes macro overlaps and optimizes macro positions and orientations effectively and efficiently for modern mixed-size circuit designs.
 
 \subsection{Classic Routing Solvers}
 \subsubsection{Global Routing Solvers}
 The global routing approaches can be divided into two types: concurrent and sequential. The sequential approach has been proved to be very effective in practice and considerably faster than the concurrent approach, but it highly relies on the ordering of the nets and thus, being prone to a sub-optimal solution. 
 
 Concurrent approaches attempt to handle numerous nets simultaneously but are typically too expensive to be applied on today’s large designs, which may contain up to a million nets. BoxRouter~\cite{cho2007boxrouter} implements progressive integer linear programming (ILP) and adaptive maze routing to effectively diffuse the congestion. However, its progressive ILP routing formulation only covers L-shape patterns, and it fails in difficult scenarios when most nets must be detoured in complex patterns. On top of it, BoxRouter 2.0~\cite{cho2007boxrouter} further provides more powerful and systematic way of eliminating congestion and assigning layers to wires. GRIP~\cite{wu2009grip,wu2010parallel} is based on a partitioning strategy in a full 3D manner and obtains the best wirelength among the open literature, but when compared to other recent global routers, GRIP requires prohibitive overall runtime. 
 
 Sequential approaches often use net decomposition~\cite{chu2007flute}, maze routing~\cite{lee1961algorithm}, pattern routing~\cite{kastner2002pattern}, or negotiation-based rip-up and rerouting (NRR), and only one net is routed at a time. 
 Archer~\cite{ozdal2009archer} explores the congestion histories, and adopts a Lagrangian relaxation-based bounded-length min-cost topology improvement algorithm that enables Steiner trees to change dynamically for congestion optimization. NTUgr~\cite{chen2009high} replaces iterative NRR by enhanced iterative forbidden-region rip-up/rerouting (IFR). FastRoute~\cite{pan2012fastroute}, on the other hand, integrates several novel techniques: fast congestion-driven via-aware Steiner tree construction, 3-bend routing, virtual capacity adjustment, multisource multi-sink maze routing, and spiral layer assignment. NTHU-Route 2.0~\cite{chang2010nthu} improves NTHU-Route~\cite{gao2008new}, an earlier version by a new history-based cost function and new ordering methods for congested region identification and rip-up and reroute. NCTU-GR 2.0~\cite{liu2013nctu} outperforms the foregoing global routers by applying two bounded-length maze routing (BLMR) algorithms (i.e. optimal-BLMR and heuristic-BLMR), a rectilinear Steiner minimum tree aware routing scheme
 , a collision-aware rip-up and rerouting scheme and a 3-D wire length optimization technique.
 
\subsubsection{Detailed Routing Solvers}
 Since the 1970s, detailed routing has been extensively researched (e.g. \cite{yoshimura1982efficient}) 
 , and rip-up and reroute, such as the one in Mighty~\cite{shin1987detailed} has been the most common technique for detailed routing. Nevertheless, when dealing with congested designs, such a sequential net-by-net method is unproductive and frequently results in unneeded detours. DUNE~\cite{cong2001dune} and MR~\cite{chang2004mr} develop multilayer techniques to handle full-chip gridless routing, in which the routing passes through a coarsening and uncoarsening phase. These multilevel routers, however, continue to use the sequential rip-up and reroute technique. Several attempts have been made to evaluate nets more concurrently during detailed routing. Based on Boolean satisfiability, \cite{nam2002new} suggests a thorough FPGA router which delivers exceedingly long runtime despite good solution quality. Track assignment is introduced in \cite{batterywala2002track} as a step between global and detailed routing. Segments taken from the global routing solution are assigned to routing tracks in track assignment. \cite{ozdal2009detailed} introduces an ingenious method for doing escape routing for dense pin clusters, which is a major bottleneck in detailed routing. However, the technique is not recommended for solving detailed routing on a whole-chip scale. Routing frameworks based on rules are well-suited to traditional design flow and have been in use for decades. To execute correct pin access, \cite{nieberg2011gridless} computes various pin access path candidates and selected the shortest paths from pins to grid points that did not violate any design criteria. RegularRoute~\cite{zhang2011regularroute} frames the global segment assignment problem inside each group of routing tracks as a maximum weighted independent set problem, then used regular routing patterns in a bottom-up layer-by-layer framework. Under self-aligned double patterning limitations, \cite{xu2016parr} suggests a pin access-driven rip-up and reroute scheme. For mixed-cell-height circuits, \cite{li2018routability} introduces a pin access-aware legalizing technique. Due to the spirited ISPD'18 and ISPD'19 routing contest \cite{mantik2018ispd,liu2019ispd}, some new work has been completed. \cite{kahng2018tritonroute} divides each layer into parallel panels and expressed the routing problem as an integer linear program on each panel. \cite{sun2018multithreaded} modifies the notion of hit points and employs through violations to assign tracks. Dr. CU~\cite{chen2019detailed} proposes an algorithm for finding the best path while keeping the minimum-area requirement in mind. The subsequent Dr. CU 2.0~\cite{li2019dr} handles hard-to-access pins and new design rules including length-dependent parallel run length spacing, end-of-line spacing with parallel edges, and corner-to-corner spacing.

\section{Machine Learning for Placement}

\subsection{Traditional Placers Enhancement}
Most traditional placers mentioned above perform heavy numerical computation for large-scale optimization problem on the CPUs, which lacks exploration of GPU’s opportunity. DREAMPlace~\cite{lin2020dreamplace} is inspired by the idea that the analytical placement problem is analogous to training a neural network. They both involve optimizing parameters and minimizing a cost function. Based on the state-of-the-art analytical placement algorithm RePlAce, DREAMPlace implements hand-optimized key operators by deep learning toolkit PyTorch and achieves over $30\times$ speedup against CPU-based tools. PL-GNN~\cite{lu2021law} presents a graph learning-based framework that provides placement guidance for commercial placers by generating cell clusters based on logical affinity information and attributes of design instances. PADE~\cite{ward2012pade} improves data path logic through automatic data path extraction and evaluation, in which the placement of data path logic is conducted separately from random logic. \cite{agnesina2020vlsi} proposes a deep reinforcement learning (RL) framework to optimize the placement parameters of commercial EDA tool. An agent learns to tune parameters autonomously, trained solely by RL from self-search. Handcrafted features along with graph embeddings generated using unsupervised Graph Neural Networks are adopted for generalization to unseen netlists.

\subsection{Placement Decision Making} 
Learning-based methods for placement decision especially RL have been proposed to obtain the generalization ability. 
Existing RL applications have demonstrated the effectiveness on macro placement, where there are typically fewer than 1000 macros to place. 
Google~\cite{mirhoseini2021graph} proposes an end-to-end learning method for macro placement that models chip placement as a sequential decision making problem. In each step, the RL agent places one macro and target metrics are used as reward until the last action. GNN is adopted in the value network to encode the netlist information and deconvolution layers in the policy network output the mask of current macro position. 
DeepPlace \cite{cheng2021joint} first proposes a joint learning technique for the placement of macros and standard cells by the integration of reinforcement learning with a gradient-based classical cell placer (\cite{lin2020dreamplace}). To further bridge the placement with the subsequent routing task, they also develop a joint learning approach DeepPR via RL to fulfill both placement and routingfor macros. Table~\ref{tab:comparison} compares  learning-based placers in terms of placement target and learning protocol.

\subsection{Prediction Model Embedded in Placement}
ML also assists placers to optimize complicated objectives like routability by embedding prediction models, as it is difficult to foresee routing congestion accurately during placement. \cite{huang2019routability} proposes the first  routability driven macro placement with deep learning. A CNN-based routability prediction model is proposed and embedded into a macro placer such that a good macro placement with minimized design rule check (DRC) violations can be derived through SA optimization process. \cite{chan2017routability} presents a learning based algorithm to predict DRC violations in detailed routing and automatically improve the routability of these designs. \cite{liu2021global} predicts congestion hotspots and then incorporates this prediction model into a placement engine, showing how an ML-based routing congestion estimator can be embedded into the global placement stage.

\subsection{Challenges and Limitations for Placement}

The major challenges for ML applications in placement lie in two folds: long feedback loop and high requirement of scalability. Placement objectives like routability cannot be evaluated until routing finished; hence, it may take hours to obtain the feedback in the optimization loop, which is unaffordable to make thousands of queries. 
Modern placers need to handle tens of thousands of macros and millions of standard cells within several hours. Such requirement of scalability is still beyond the capability of existing ML approaches.

\section{Machine Learning for Routing}

\subsection{Learning-aided Routability Prediction}
  In the placement step, the essential requirements of routing design guidelines must be considered. However, it is difficult to precisely and quickly estimate routing information during the placement step, and researchers have lately used machine learning to overcome this problem. 
  Table~\ref{tab:ml4cong} summarize the recent efforts on routability prediction, which can be categorized into congestion count prediction and congestion location prediction at different design stages. 
Congestion count denotes congestion related metrics such as total congestion and number of design rule violations, 
while congestion locations require detailed locations of congestion or design rule violations, usually represented as a 2D map. 

\input{ml4cong.tex}


\textbf{Congestion count.} As mentioned above, it denotes the overall amount of routing congestion, useful to evaluate how good a placement solution is. Efficient prediction of congestion count can reduce the turn-around time in the design flow by avoiding the time-consuming routing stage. 
\cite{qi2014accurate,zhou2019supervised} capture multiple factors in global routing and enable prediction of detailed routing congestion using multivariate adaptive regression splines (MARS). \cite{tabrizi2018machine,maarouf2018machine} attempt to predict global routing congestion at placement stage with linear regression (LR), random forest (RF), and MLP models for datasets from ASIC and FPGA. Other studies like \cite{zhou2015accurate,chan2016beol} aim at predicting the congestion count at detailed routing given cell placement, 
and \cite{cheng2018evaluation} tries to predict the congestion count at global routing given only macro placement.  

  \textbf{Congestion location.} Its accurate prediction is necessary to effectively guide the placement and routing optimization, as it can help reserve enough space for congested regions. 
  RouteNet~\cite{xie2018routenet} is the first attempt to utilize CNN to forecast the locations of design rule checking (DRC) hotspots given cell placement and global routing information. 
  A customized fully convolutional network (FCN) is constructed taking features like rectangle uniform wire density (RUDY), as a pre-routing congestion estimator, and global routing congestion map. 
  Predicting global routing congestion locations at placement is helpful to guide routability optimization in early stages. 
  Thus, PROS~\cite{chen2020pros}, \cite{pui2017clock}, \cite{yu2019painting}, \cite{alawieh2020high}, and DLRoute~\cite{al2021deep} attempt to learn the correlation between congestion locations at global routing and cell placement. 
  As both the features and labels can be represented as image-like tensors, many studies transform the problem into image generation tasks and leverage FCN and conditional GAN to build the correlation. 
  \cite{liang2020drc} moves one step further to directly predict congestion locations at detailed routing (i.e., locations of design rule violations) from cell placement. A customized CNN architecture, J-Net (an extension of U-Net architecture), is proposed as opposed to a plug-in use of machine learning modules. This work converts the density of pins and macros in placement results into images and optimizes an encoder-decoder model using a pixel-wise loss function. The network outputs a heat map, showing where detailed routing congestion might occur.

Among learning-aided prediction tasks, the more stages to skip in Fig.~\ref{fig:flow}, the more difficult the tasks are, as we need to build models correlating with more stages. Thus, the accuracy requirement varies from task to task. Predicting congestion locations at detailed routing at early stages like cell placement or even macro placement can expedite the design.

\begin{table}[tb!]
\centering
\resizebox{.49\textwidth}{!}{
\begin{tabular}{|c||cc||cc|}
\hline
         & \multicolumn{2}{c||}{\textbf{Target for placement}}  & \multicolumn{2}{c|}{\textbf{Learning protocol}}                                      \\[0.2ex] \hline
\textbf{Publication} & \multicolumn{1}{c|}{\textbf{Macro}} & \textbf{Standard cell} & \multicolumn{1}{c|}{\textbf{Module}} & \multicolumn{1}{c|}{\textbf{Reward using}}   \\ [0.3ex] \hline
\cite{cheng2021joint}         & \multicolumn{1}{c|}{RL}      &    DNN backprop            & \multicolumn{1}{c|}{CNN+GNN}       & \multicolumn{1}{c|}{cell placement}             \\ \hline
\cite{mirhoseini2021graph}         & \multicolumn{1}{c|}{RL}      &     NA          & \multicolumn{1}{c|}{GNN}  &      \multicolumn{1}{c|}{ macro placement}                  \\ \hline
\cite{vashisht2020placement}         & \multicolumn{1}{c|}{RL}      &     NA         & \multicolumn{1}{c|}{MLP}       & \multicolumn{1}{c|}{ macro placement}                     \\ \hline
\cite{he2020learn}         & \multicolumn{1}{c|}{RL}      &     NA          & \multicolumn{1}{c|}{MLP}       & \multicolumn{1}{c|}{ macro placement}                     \\ \hline
\cite{lin2020dreamplace}         & \multicolumn{1}{c|}{	heuristic}      &    DNN backprop            & \multicolumn{1}{c|}{NA}       & \multicolumn{1}{c|}{NA}                     \\ \hline
\end{tabular}}
\vspace{-6pt}
\caption{Comparison of learning-based placers. RL is the most popular paradigm for macro placement, while learning for standard cell placement still relies on traditional placers for further enhancement. Note that \protect\cite{cheng2021joint} is a joint learning approach for solving placement of macro and standard cells, whereby the two kinds of components are sequentially arranged by reinforcement
learning and neural network formed gradient optimization respectively and reward is based on the full placement result. \protect\cite{lin2020dreamplace} is not a RL method, thus there is no reward function.}
\label{tab:comparison}
\end{table}
 
\subsection{Deep Neural Networks for Routing}
  Rather than the methods mentioned above predicting the congestion information of placement, some works employ DNNs to directly handle routing problems, with a little or even without the assistance of traditional routing techniques. \cite{jain2017training} presents a fully convolutional neural network learning to route a circuit layout with appropriate choices of metal tracks and wire class combinations. Encoded layouts containing spatial location of pins to be routed are fed into the network, and after 15 fully convolutional layers followed by a comparator, 8 layout layers are produced, which are then decoded to obtain the routed layouts. This work formulates routing as a binary segmentation problem on a per-pixel per-layer basis, where the network is trained to correctly classify pixels in each layout layer to be \textit{on} or \textit{off}. \cite{he2020circuit} models the circuit routing as a sequential decision-making problem, and solve it by Monte Carlo tree search (MCTS) with DNN guided rollout. A recent study~\cite{utyamishev2020late} proposes a global router that learns from routed circuits and autonomously routes unseen layouts. Different from traditional routing flow, this approach redefines the global routing as a classical image-to-image processing problem and handles the imaging problem in a unified, single-step non-iterative manner with a deep learning system, comprising a variational autoencoder and a custom loss.

\subsection{Reinforcement Learning for Routing}
RL is also a promising way to tackle routing, as it can be seen as a process that comprises decision-making phases. A DQN agent~\cite{liao2020deep}, as one of the first attempts to combine RL with global routing, learns to decide the routing direction on a 3D grid graph at each step, e.g. traveling north, south, and so on. For detailed routing, \cite{liao2020attention} presents an attention-based REINFORCE method for obtaining routing orders, followed by a classical pattern router to finish the routing given the order, for small benchmarks with up to thousands nets. \cite{ROUTE_TCAD2021_Lin} tackles routing ordering with an asynchronous actor-critic framework for routing millions of nets and improves the solution quality over the state-of-the-art detailed router with policy distillation. \cite{ren2021standard} employs a genetic algorithm to generate initial routing choices and then uses RL to progressively handle design rule violations in standard cell routing. Regarding rectilinear Steiner minimum tree (RSMT) construction~\cite{hartmanis1982computers}, 
a process that is a fundamental problem in EDA and computer science and typically runs millions of times in traditional global routers, REST~\cite{liu2021rest} is the first attempt to solve RSMT construction using a machine learning-based method. A new concept, rectilinear edge sequence (RES), is proposed to encode an RSMT solution
, and an actor-critic model
is devised to construct an RSMT.

\subsection{Challenges and Limitations for Routing}
Despite the existing efforts on learning-based routing, it is difficult for current techniques to systematically outperform classical routing algorithms in both efficiency and solution quality under fair comparison. Most learning-based techniques work well on small circuits with thousands of nets, while a practical routing engine needs to handle millions of nets on an ultra-large 3D grid graph ($ > 1000 \times 1000 \times 10$) efficiently and produce high-quality solutions. 
 
\section{Open-source and Public Datasets}


\paragraph{Open-source (OS).} This area is becoming more and more open, and we list a few representative OS examples. Google releases a framework for floorplanning with distributed DRL\footnote{https://github.com/google-research/circuit\_training}. The end-to-end learning approach DeepPlace for macros and standard cells placement problem is also publicly available\footnote{https://github.com/Thinklab-SJTU/EDA-AI}. Another project DREAMPlace provides a deep learning toolkit-enabled VLSI placement tool\footnote{https://github.com/limbo018/DREAMPlace}. 
The attempt to combine DRL with global routing is now available\footnote{https://github.com/haiguanl/DQN\_GlobalRouting}.
We believe that these open-source resources would foster collaborations between academic and industry placement and routing tools.

\paragraph{Datasets.} Starting from the pure wirelength-driven formulations in 2005, regular research contests for placement are held every year at ISPD and ICCAD to provide various circuit designs contributed by industry in the common Bookshelf format. From then on, the contests broadened their specific optimization objectives toward cell density, global routability~\cite{viswanathan2011ispd}, detailed routability~\cite{yutsis2014ispd}, and timing~\cite{kim2015iccad}. A popular benchmark for both placement and routing is the ISPD 2015~\cite{bustany2015ispd}. It consists of five circuits, and each of them has multiple floorplans produced from setting different macro locations with predefined heuristics. ISPD announced two global routing contests in 2007~\cite{nam2007ispd} and 2008~\cite{nam2008ispd} and two initial detailed routing contests in 2018~\cite{mantik2018ispd} and 2019~\cite{liu2019ispd}, respectively, along with the benchmarks.

\section{Conclusion and Outlook}
In modern physical design flow, it takes human experts weeks to iterate the placement tools in order to produce solutions with no design rule check violations after routing stage. With rapid development of machine learning, a promising solution for this obstacle is to propose efficient and effective learning framework for solving placement and routing either sequentially or concurrently. \cite{cheng2021joint} designs a joint learning approach for either macro placement and routing or placement of macros and standard cells. However, this is not a complete design cycle since millions of or even billions of standard cells are left behind. For future works, learning-based routing solvers merit particular attention as the final building block for the entire end-to-end learning paradigm.

\small
\bibliographystyle{named}
\bibliography{place_routing}

\end{document}

%% file: flow.tex
\begin{figure}[tb!]
\centering
\begin{tikzpicture}[thick, scale=0.8, every node/.style={transform shape}]
\node[rectangle, draw, text centered, text width=4.5em](macropl){Macro Placement}; 
\node[rectangle, draw, text centered, text width=4.5em, right=1em of macropl](gp){Global Placement};
\node[rectangle, draw, text centered, text width=4.5em, right=1em of gp](dp){Detailed Placement};
\node[rectangle, draw, text centered, text width=4.5em, right=1em of dp](gr){Global Routing};
\node[rectangle, draw, text centered, text width=4.5em, right=1em of gr](dr){Detailed Routing};

\path[draw, -latex](macropl) -- (gp);
\path[draw, -latex](gp) -- (dp);
\path[draw, -latex](dp) -- (gr); 
\path[draw, -latex](gr) -- (dr); 

\draw [
    thick,
    decoration={
        brace,
        raise=0.45cm
    },
    decorate
] (gp.west) -- (dp.east) 
node [pos=0.5,anchor=south,yshift=0.65cm] {Cell Placement};

\draw [
    thick,
    decoration={
        brace,
        mirror,
        raise=0.45cm
    },
    decorate
] (macropl.west) -- (dp.east) 
node [pos=0.5,anchor=north,yshift=-0.65cm] {Placement}; 

\draw [
    thick,
    decoration={
        brace,
        mirror,
        raise=0.45cm
    },
    decorate
] (gr.west) -- (dr.east) 
node [pos=0.5,anchor=north,yshift=-0.65cm] {Routing}; 
\end{tikzpicture}
\vspace{-20pt}
\caption{
Example of a simplified placement and routing flow. 
}
\label{fig:flow}
\end{figure}
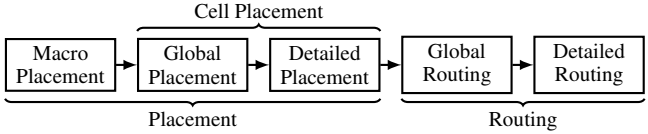

%% file: ml4cong.tex
\begin{table*}[tbh]
\centering
\resizebox{.99\textwidth}{!}{
\begin{tabular}{|c|c|c||c|c|c|}
\hline
\textbf{Task type}                                                                                                & \textbf{Feature from}                    & \textbf{Label at}                        & \textbf{Publication}                     & \textbf{Backbone models}           & \textbf{Benchmark}             \\ \hline
\multirow{7}{*}{\begin{tabular}[c]{@{}c@{}}Congestion Count\\ (Scalar or Vector as Labels)\end{tabular}} & \multirow{2}{*}{Global Routing} & \multirow{2}{*}{Detailed Routing} & \cite{qi2014accurate}           & MARS                      & \multirow{2}{*}{ASIC} \\ \cline{4-5}
                                                                                                         &                                 &                                   & \cite{zhou2019supervised}       & MARS                      &                       \\ \cline{2-6} 
                                                                                                         & \multirow{2}{*}{Cell Placement} & \multirow{2}{*}{Global Routing}   & \cite{tabrizi2018machine}       & MLP                       & ASIC                  \\ \cline{4-6} 
                                                                                                         &                                 &                                   & \cite{maarouf2018machine}       & LR, RF, MLP           & FPGA                  \\ \cline{2-6} 
                                                                                                         & \multirow{2}{*}{Cell Placement} & \multirow{2}{*}{Detailed Routing} & \cite{zhou2015accurate}         & MARS                      & \multirow{2}{*}{ASIC} \\ \cline{4-5}
                                                                                                         &                                 &                                   & \cite{chan2016beol}             & MARS, SVM                 &                       \\ \cline{2-6} 
                                                                                                         & Macro Placement                 & Global Routing                    & \cite{cheng2018evaluation}      & LR, RF, Boosting, MLP & ASIC                  \\ \hline
\hline
\multirow{7}{*}{\begin{tabular}[c]{@{}c@{}}Congestion Locations\\ (2D Map as Labels)\end{tabular}}       & Global Routing                  & Detailed Routing                  & RouteNet \cite{xie2018routenet} & FCN                       & ASIC                  \\ \cline{2-6} 
                                                                                                         & \multirow{5}{*}{Cell Placement} & \multirow{5}{*}{Global Routing}   & PROS \cite{chen2020pros}        & FCN                       & ASIC                  \\ \cline{4-6} 
                                                                                                         &                                 &                                   & \cite{pui2017clock}             & LR, SVM               & \multirow{4}{*}{FPGA} \\ \cline{4-5}
                                                                                                         &                                 &                                   & \cite{yu2019painting}           & Conditional GAN           &                       \\ \cline{4-5}
                                                                                                         &                                 &                                   & \cite{alawieh2020high}          & Conditional GAN           &                       \\ \cline{4-5}
                                                                                                         &                                 &                                   & DLRoute \cite{al2021deep}       & CNN                       &                       \\ \cline{2-6} 
                                                                                                         & Cell Placement                  & Detailed Routing                  & J-Net \cite{liang2020drc}       & U-Net                     & ASIC                  \\ \hline
\end{tabular}
}
\vspace{-6pt}
\caption{
Summary of recent publications on learning-aided routability prediction in terms of congestion count and locations. 
}
\label{tab:ml4cong}
\end{table*}